%% file: eacl2023.tex
\pdfoutput=1
\documentclass[11pt]{article}
\usepackage[svgnames,usenames,dvipsnames]{xcolor}

\usepackage[]{EACL2023}

\usepackage{times}
\usepackage{latexsym}

\usepackage[T1]{fontenc}
\usepackage[utf8]{inputenc}

\usepackage{microtype}

\usepackage{inconsolata}

\usepackage[ruled,algo2e]{algorithm2e}
\usepackage{algorithm}
\SetKwRepeat{Do}{do}{while}
\input{math_commands}

\usepackage{makecell}
\usepackage{booktabs}   % professional-quality tables
\usepackage{amssymb}
\usepackage{dsfont}
\usepackage{comment}
\usepackage{url}
\usepackage{hyperref}
\hypersetup{
    colorlinks= true,
    citecolor = DarkBlue,
    linkcolor= DarkBlue,
    urlcolor= DarkBlue,
}

\title{Reliable Gradient-free and Likelihood-free Prompt Tuning}

\author{
    Maohao Shen\textsuperscript{\rm 1},
    \bf Soumya Ghosh\textsuperscript{\rm 3}, 
    \bf Prasanna Sattigeri\textsuperscript{\rm 3},\\
     \bf Subhro Das\textsuperscript{\rm 3},
    Yuheng Bu\textsuperscript{\rm 2},
    \bf Gregory Wornell\textsuperscript{\rm 1}\\
    \textsuperscript{\rm 1} Massachusetts Institute of Technology\\
    \textsuperscript{\rm 2} University of Florida\\
    \textsuperscript{\rm 3} MIT-IBM Watson AI Lab, IBM Research\\
}

\begin{document}
\maketitle

\input{Sections/Abstract}
\input{Sections/Introduction}

\input{Sections/Background}

\input{Sections/Problem}

\input{Sections/Method}
\input{Sections/Exp}
\input{Sections/conclusion}

\section{Limitations}
We explored methods for learning a distribution over prompts for tuning PLMs with only API access. We rely on approximate inference algorithms to infer these distributions. Since the true posterior is intractable, effectively evaluating the quality of the inferred approximate posteriors is challenging. Here, we use downstream metrics to compare different algorithms. However, such metrics conflate the quality of posterior approximation with predictive performance. Assessing the quality of approximate posteriors remains an open problem. Another limitation of our ABC-based approach is that it is more expensive than approaches that can exploit gradient information. Improving the computational efficiency of such approaches comprises interesting future work.
% : the \textit{Gradient-free} and \textit{likelihood-free} prompt tuning, a performance gap remains between such new setting and the \textit{gradient-free} setting. The classical SBI approaches like the ABC-SMC algorithm utilized in this work may not be optimal SBI algorithm for prompt tuning problems. It is worth investigating more state-of-art SBI algorithms. While the key idea of this work is to infer the posterior distribution over prompts through approximate inference, evaluating the quality of the approximate posteriors remains an open challenge. Meanwhile, the training time of our proposed method is longer than the existing approaches, and improving the computational efficiency can also be a future work. 

\section{Acknowledgements}
This work was supported, in part, by the MIT-IBM Watson AI Lab under Agreement No.~W1771646, and NSF under Grant No.~CCF-1816209.

\bibliography{refs}
\bibliographystyle{acl_natbib}

%Appendix
\clearpage
\appendix
\input{Sections/Appendix}
\end{document}

%% file: math_commands.tex
\usepackage{amsmath,amsfonts,bm, amsthm, mathrsfs, mathtools}

\numberwithin{theorem}{section}
\numberwithin{lemma}{section}
\numberwithin{proposition}{section}
\numberwithin{corollary}{section}

\def\eqref#1{equation~\ref{#1}}
\def\1{\bm{1}}

\def\vx{{\bm{x}}}

\def\vz{{\bm{z}}}

\DeclareMathAlphabet{\mathsfit}{\encodingdefault}{\sfdefault}{m}{sl}
\SetMathAlphabet{\mathsfit}{bold}{\encodingdefault}{\sfdefault}{bx}{n}

\def\gD{{\mathcal{D}}}

\def\gH{{\mathcal{H}}}

\def\gL{{\mathcal{L}}}

\def\gN{{\mathcal{N}}}

\def\sR{{\mathbb{R}}}

\DeclareMathOperator*{\argmax}{arg\,max}
\DeclareMathOperator*{\argmin}{arg\,min}

%% file: Sections/Abstract.tex
\begin{abstract}
Due to privacy or commercial constraints, large pre-trained language models (PLMs) are often offered as black-box APIs. Fine-tuning such models to downstream tasks is challenging because one can neither access the model's internal representations nor propagate gradients through it. This paper addresses these challenges by developing techniques for adapting PLMs with only API access. Building on recent work on soft prompt tuning, we develop methods to tune the soft prompts without requiring gradient computation. Further, we develop extensions that in addition to not requiring gradients also do not need to access \emph{any} internal representation of the PLM beyond the input embeddings. Moreover, instead of learning a single prompt, our methods learn a distribution over prompts allowing us to quantify predictive uncertainty. Ours is the first work to consider uncertainty in prompts when only having API access to the PLM. Finally, through extensive experiments, we carefully vet the proposed methods and find them competitive with (and sometimes even improving on) gradient-based approaches with full access to the PLM.

\end{abstract}

%% file: Sections/Introduction.tex
\section{Introduction}
Pre-trained language models (PLMs) are versatile learners and demonstrate impressive few-shot capabilities \cite{ brown2020language} and promising performance~\citep{radford2018improving, devlin2018bert, raffel2020exploring, lewis2019bart} on various downstream tasks such as  text classification~\citep{kowsari2019text}, commonsense reasoning~\citep{zellers2018swag}, question answering~\citep{rajpurkar2016squad}, and machine translation~\citep{bahdanau2014neural}. 
%\ps{{Cite and list some recent paper and downstream tasks}}.

The conventional approach to adapting PLMs to downstream tasks involves fine-tuning the model~\citep{Peters2018DeepCW, devlin2018bert}. Although fine-tuning is effective, it can be challenging to do in practice. First, fine-tuning large language models are compute and memory intensive, e.g., a large model like GPT-3~\citep{brown2020language} contains billions of parameters. Further, it is inefficient to adapt a PLM to a large number of downstream tasks since each task would require storing a copy of model parameters.

\emph{Prompt tuning} alleviates these issues by providing an efficient way to adapt a PLM to a downstream task. It only learns a small number of prompt parameters while keeping the large PLM frozen but still achieves comparable performance to fine-tuning the entire PLM~\cite{liu2021pre, shin2020autoprompt, lester2021power, liu2021gpt}.

% as fine-tuning approaches

%\ps{Talk about benefits over traditional fine-tuning and cite related works.}

Although more efficient than traditional fine-tuning, prompt tuning still requires the propagation of gradients through the entire PLM. Beyond being computationally expensive, this may not be possible due to privacy risks or legal and commercial constraints. In fact, large PLMs are often only made available in the form of black-box APIs~\citep{brown2020language}. Motivated by these observations, a recent line of research~\citep{sun2022black, sun2022bbtv2} has started exploring \emph{gradient-free} approaches to prompt tuning. BBT~\citep{sun2022black} optimizes continuous prompt by leveraging the derivative-free optimization algorithms, and BBTv2~\citep{sun2022bbtv2} improves over BBT by optimizing multiple deep prompts at various intermediate layers of PLM. Although these approaches are \textit{gradient-free}, they still assume that intermediate layers of the model being tuned are accessible. % prompt tuning is still not fully exploited, and some practical concerns exist when applying the existing methods. 

Moreover, when deploying an NLP model in a real-world setting, it is inevitable to encounter unexpected scenarios. For example, the test data to be predicted might originate from out-of-distribution resources~\citep{arora2021types}. For the model to be useful in such scenarios, it is essential that the model is able to quantify the uncertainty associated with its predictions and that these uncertainties are well-calibrated.
%A reliable model should successfully identify such samples and prevent taking the risk of making wrong predictions. Second, the aforementioned prior works cannot directly be applied to practical settings since they still assume that the \textit{logits} of PLM are accessible. Such an assumption is not valid for black-box APIs~\citep{brown2020language} of large language models, whose only accessible outputs are the discrete outcome labels.

%\ps{Cite  more on why do in a black-box manner?}

To this end, here we further push the limits of \textit{gradient-free} prompt tuning in two aspects:

\begin{itemize}
    \item First, we develop methods that add a layer of uncertainty quantification (UQ) aimed toward more reliable prompt tuning. We show that this improves calibration and UQ performance on several tasks, including selective classification and text Out-of-Distribution (OOD) detection.
    
    \item Second, we consider a much stricter notion of black-box setting, i.e., \textit{likelihood-free} setting, where the PLM-based API does not provide probability scores or \textit{logits} as the output, but only the discrete outcome labels. We propose a simulation-based-inference approach that yields competitive performance in the stricter setting even compared to the SOTA prior works on the relaxed black-box setting.
\end{itemize}

%% file: Sections/Background.tex
\section{Background}

\paragraph{Prompt Tuning}
Prompting, in the simplest form, involves appending manually curated words or tokens to a text input such that the language model, conditioned on such an augmented input, generates the desired output \cite{liu2021pre}. Such curated prompts were shown to be much more efficient than fine-tuning the entire PLM \cite{brown2020language}. However, curating good prompts for a new task can be difficult without deep domain expertise \cite{liu2021gpt,zhao2021calibrate}. One solution is to search the space of discrete prompts\cite{shin2020autoprompt,gao2020making}. This search in discrete space can be a hard optimization problem. Recent works instead learn continuous or soft prompts in the form of a small number of free parameters injected into certain layers of the PLM \cite{li2021prefix}. In this paper, we work with the simpler form of continuous \textit{prompt tuning}, where the free parameters are only injected in the embedding layer \cite{lester2021power}.

\paragraph{Gradient-free Prompt Tuning}
\textit{Gradient-free} prompt tuning aims to learn the continuous prompt without the propagating gradients through the PLM. BBT \citep{sun2022black} utilizes derivative-free optimization algorithms to optimize the continuous prompt. BBTv2~\citep{sun2022bbtv2} extends BBT by incorporating the idea of \emph{deep prompt tuning}, which optimizes the deep prompt injected at additional intermediate layers of the PLM. Since our goal is to treat the PLM as a black-box, deep prompt tuning is out of the scope of this work. We instead focus on the problem setting of the original BBT~\citep{sun2022black} that learns a single prompt at the input layer.

\paragraph{Beyond point-estimates of prompts}
Many applications demand accurate quantification of uncertainty in predictions. This can be achieved in the prompt-tuning setting by not just learning a point estimate of the prompts but also inferring a distribution over the prompts for a given downstream task. In a non-black-box setting, to infer such a distribution, we can apply classical frequentist or Bayesian approaches. 
%statistical approaches that include frequentist methods and Bayesian methods such as ensembles, Markov Chain Monte Carlo~\citep{neal1993probabilistic, neal2011mcmc}, variational inference~\citep{wainwright2008graphical}, etc.
Although a few recent works focus on uncertainty quantification in NLP applications~\citep{arora2021types, xiao2019quantifying, desai2020calibration, kumar2019calibration}, quantifying uncertainty in prompt-tuned large language models remains a severely under explored area. Our paper is the first to explore prompt uncertainty in gradient-free settings.

\paragraph{Simulation-based Inference }
Classic approaches for statistical inference mentioned above are intractable when the likelihood function is not accessible. The problem of inferring parameters of such a black-box model, called Simulation-based Inference (SBI)~\citep{cranmer2020frontier}, is gaining popularity. Traditional SBI approaches include Approximate Bayesian
Computation (ABC)~\citep{beaumont2002approximate, marjoram2003markov, marin2012approximate, beaumont2009adaptive, bonassi2015sequential} and synthetic likelihood (SL)~\citep{wood2010statistical, turner2014generalized}. 
% but they usually suffer from computational efficiency and heuristic approximation issues. 
More recently, the neural density estimation-based approaches utilize the powerful deep neural network density estimator to directly learn the likelihood, i.e., Sequential Neural Likelihood Estimation (SNLE)~\citep{lueckmann2017flexible, greenberg2019automatic}, or the likelihood ratio, i.e., Sequential Neural Ratio Estimation (SNRE)~\citep{papamakarios2019sequential}, or the posterior, i.e., Sequential Neural Posterior Estimation (SNPE)~\citep{hermans2020likelihood, durkan2020contrastive}.

%\paragraph{Uncertainty Quantification}
%Uncertainty Quantification (UQ) aims to construct robust machine learning models by quantitatively measuring the reliability of model’s prediction. Typical UQ approaches include two main categories: \textit{intrinsic} approaches that provide uncertainty estimation along with the model prediction, including neural networks with homoscedastic/heteroscedastic noise models \citep{wakefield2013bayesian} and quantile regression \cite{koenker1978regression}, and Bayesian techniques such as Bayesian neural networks (BNNs) \cite{neal2012bayesian, blundell2015weight, welling2011bayesian}, Gaussian processes \cite{gpbook} and ensemble methods~\citep{lakshminarayanan2017simple};  \textit{extrinsic} methods that extract uncertainties in a post-hoc manner including calibration methods~\citep{kull2019beyond, guo2017calibration}, and meta-modeling approaches~\citep{chen2019confidence, jain2021deup}.

%Despite general methodologies, a few recent works focus on UQ in NLP applications~\citep{arora2021types, xiao2019quantifying, desai2020calibration, kumar2019calibration}. However, none of these works studied the NLP UQ problem from prompt tuning perspectives.

%% file: Sections/Problem.tex
\section{Problem Formulation} \label{sec:problem}
\begin{figure*}[t]
  \centering
  \includegraphics[width=0.95\linewidth]{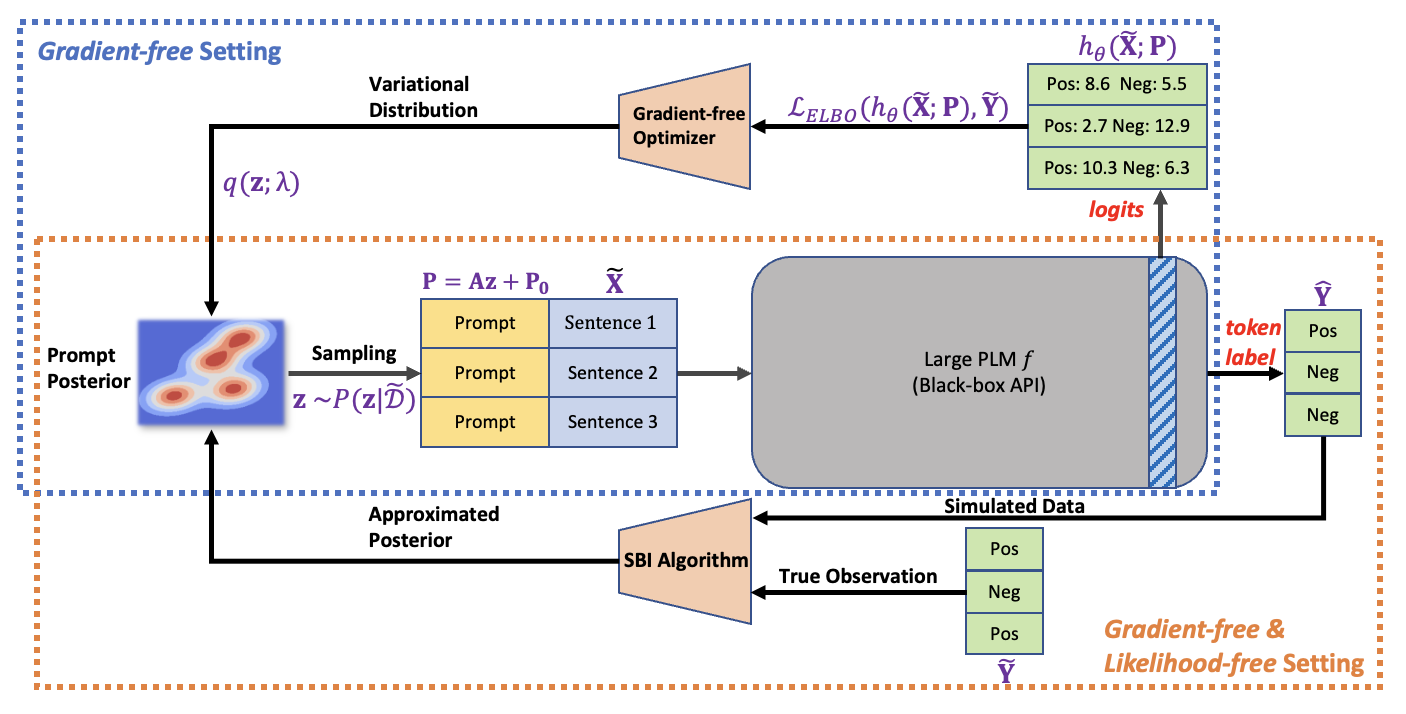}
  \vspace{-0.5em}
  \captionsetup{font=small}
  \caption{Our general goal is to estimate the posterior distribution of prompts. The \textcolor{RoyalBlue}{\textit{\textbf{Gradient-free setting}}} uses the internal \textit{logits} of PLM for optimization. Our proposed Gradient-free Variational inference approach utilizes the likelihood to compute the ELBO objective and leverage the gradient-free optimizer to optimize the variational distribution. The \textcolor{Bittersweet}{\textit{\textbf{Gradient-free and likelihood-free setting}}} can be formulated as an SBI problem, where the PLM is treated as a black-box simulator, and its output discrete outcome labels are the simulated data. The posterior samples can be efficiently approximated by the proposed ABC-SMC algorithm.}
  \label{fig:figure}
  \vspace{-0.5em}
\end{figure*}

In this paper, we focus on text classification and restrict ourselves to the few-shot learning setting considered in BBT~\citep{sun2022black}.
% We focus on the text classification task under the few-shot learning setting adopted by BBT~\citep{sun2022black}. 
Given a dataset $\gD = (\mathbf{X},\mathbf{Y}) = \{(\vx_i, y_i)\}_{i=1}^{N}$ and a pre-trained language model (PLM) $f$, we aim to adapt $f$ to predict the label $y_*$ for an unseen text passage $\vx_*$. 
We formulate the classification task as a masked language modeling problem, where the input text $\vx_i$ is converted into $\tilde{\vx}_i$ via predefined templates, e.g., adding trigger words like ``It was [MASK]'', and the labels $y_i$ are mapped to label tokens $\tilde{y}_i$ in the vocabulary such as ``great'' or ``bad''. We denote this transformed dataset $\tilde{\gD} = (\tilde{\mathbf{X}}, \tilde{\mathbf{Y}})$.
 
We use soft prompt tuning~\citep{lester2021power} to adapt $f$, i.e., we construct a continuous prompt embedding $\mathbf{P} \in \sR^{D}$ and feed it along with the converted input text $\tilde{\vx}_i$ to the PLM $f$ to generate a label token, $\hat{y}_i = f(\tilde{\vx}_i; \mathbf{P})$, where the notation  $\hat{y}_i = f(\tilde{\vx}_i; \mathbf{P})$ is short hand for $\hat{y}_i \sim \text{Cat}(\sigma(h_\theta(\tilde{\vx}_i; \mathbf{P})))$. Here, $\text{Cat}$ denotes the Categorical distribution, $\sigma$ is the softmax function, and $\theta$ represents the frozen parameters of the PLM. We use $h_\theta$ to denote all but the final layer of the PLM $f$.  Finally, we aim to learn an optimal prompt 
\begin{equation}\mathbf{P}^* = \argmin_{\mathbf{P}} -\sum_{i=1}^N \text{log Cat}(\tilde{y}_i \vert \sigma(h_\theta(\tilde{\vx}_i; \mathbf{P}))).
\label{eq:mle}
\end{equation}
This is just the standard cross-entropy loss and can be easily minimized using standard stochastic gradient based approaches provided (i) we can propagate gradients through the PLM $f$, and (ii) we can access the PLM's logits, i.e., $h_\theta(\tilde{\vx}_i; \mathbf{P})$. The problem becomes substantially more challenging when these requirements are not satisfied. 
 
When we are unable to propagate gradients through $f$, we need to rely on \textit{gradient-free} approaches to optimize Equation~\ref{eq:mle}. Recent work~\cite{sun2022black} has demonstrated promising gradient-free prompt tuning results by first employing a lower dimensional re-parameterization, $\vz \in \sR^d$ with $d<<D$, $\mathbf{P} = \mathbf{A}\vz+\mathbf{P_0}$, where $\mathbf{A} \in \sR^{D \times d}$ is a random projection matrix and $\mathbf{P_0}$ is a fixed prompt embedding, and then using gradient-free evolutionary algorithms, in particular, Covariance Matrix Adaptation Evolution Strategy (CMA-ES)~\citep{hansen2001completely,hansen2003reducing} to optimize,
\begin{align}
&\vz^* \nonumber \\
&= \argmin_{\vz} -\sum_{i=1}^N \text{log Cat}(\tilde{y}_i  \vert \sigma(h_\theta( \tilde{\vx}_i; \mathbf{A}\vz+\mathbf{P_0}))
\label{eq:pred1}
\end{align}
Going forward, we also adopt this lower dimensional parameterization, but instead of learning a point estimate $\vz^*$, we learn a distribution $p(\vz \vert \tilde{\gD})$ in a \textit{gradient-free} setting. Similar to the point estimated variants, our algorithms to learn $p(\vz \vert \tilde{\gD})$ also rely on CMA-ES. 

Next, we consider the fully black-box setting --- \emph{likelihood-free} and \emph{gradient-free}. Here, beyond being unable to propagate gradients through $f$, we are further handicapped by only observing the predicted label tokens, $\hat{y}_i = f(\tilde{\vx}_i; \mathbf{P})$ for each training instance $\tilde{\vx}_i$, and not the corresponding logits, i.e., $h_\theta(\tilde{\vx}_i; \mathbf{P})$. In this more challenging setting we found CMA-ES based approaches to be unreliable, often getting stuck in poor optima. Instead, we found it effective to pose the \textit{likelihood-free} and \emph{gradient-free} prompt tuning task as a simulation-based inference (SBI)~\citep{cranmer2020frontier} problem. We view the PLM $f$ as a black-box simulator that given a realization of $\vz$ and the text $\tilde{\vx}_i$ produces $\hat{y}_i$. We then use a sequential Monte-Carlo approximate Bayesian computation (SMC-ABC) approach to infer the distribution $p(\vz \vert \tilde{\gD})$. 

Finally, we use the distribution $p(\vz \vert \tilde{\gD})$ to characterize the uncertainty in predictions via the predictive distribution  $p(\tilde{y}\vert \tilde{\vx}, \tilde{\gD}) = \int p(\tilde{y}\vert\tilde{\vx}; \vz) p(\vz\vert\tilde{\gD}) d\vz$. We form Monte-Carlo approximations to this integral. In the \emph{gradient-free} case, this is, $$p(\tilde{y}\vert \tilde{\vx}, \tilde{\gD}) \approx \frac{1}{S} \sum_{s=1}^S p(\tilde{y}\vert\tilde{\vx}; \vz_s),$$ where $\vz_s \sim p(\vz\vert\tilde{\gD})$. In the \emph{likelihood-free} and \emph{gradient-free} case, since we only have access to the label tokens, we approximate the predictive distribution, 
\begin{equation}
p(\tilde{y}=c|\tilde{\vx}; \tilde{\gD}) \approx \frac{1}{S} \sum_{s=1}^{S} \mathds{1}\{\hat{y}_s=c \}, 
\label{eq:pred2}
\end{equation}
where $\hat{y}_s = f(\vx; \mathbf{A}\vz_s + \mathbf{P}_0)$, $\vz_s \sim p(\vz \vert \tilde{\gD})$.
In Section~\ref{sec:exp} we empirically demonstrate that by characterizing the uncertainty in $\vz$ through $p(\vz \vert \tilde{\gD})$ we get better calibrated predictive uncertainties, improved selective classification, and out-of-distribution detection.

%% file: Sections/Method.tex
\section{Methods}

We now describe our methods in greater detail. First, we discuss two algorithms for the \textit{gradient-free} setting in~\ref{sec:Ensembles} and \ref{sec:ELBO}. After that, we focus on addressing the \textit{gradient-free} and \textit{likelihood-free} setting from the SBI perspective in~\ref{sec:SBI}.

\subsection{Prompt Ensembles} \label{sec:Ensembles}
%One efficient way to estimate the predictive label distribution is approximating it by finite number of Monte Carlo samples, i.e., $P(\tilde{y}|\tilde{\vx}; \gD) \approx \frac 1 K \sum_{k=1}^K P(\tilde{y}|\tilde{\vx}; \vz_k)$. 
Deep ensembles~\citep{lakshminarayanan2017simple} are a simple yet effective technique for quantifying uncertainty in deep neural network predictions. They generate a uniformly-weighted ensemble by re-training the same neural network from different random initialization. Leveraging the CMA-ES algorithm~\citep{hansen2001completely,hansen2003reducing}, we can adapt this idea to \textit{gradient-free} prompt tuning.

CMA-ES is an evolutionary strategy that maintains a multivariate normal distribution $\gN(m_t,\sigma^2_t C_t)$ over a population of solutions. Each iteration of the algorithm involves sampling a set of possible solutions and updating the normal distribution to favor low loss solutions. To build a prompt ensemble, we run $S$ instances of CMA-ES, each initialized with a different random initialization of the mean $m_t$ and variance $\sigma^2_t$ and record the optimized prompt embeddings produced by each instance. This collection of $S$ prompt embeddings $\{\vz_s\}_{s=1}^S$ form the distribution $p(\vz\vert\tilde{\gD})$ and are used to approximate the predictive distribution via Equation~\ref{eq:pred1}.

\subsection{Gradient-free Variational Inference} \label{sec:ELBO}
An alternative way to estimate the predictive distribution is by approximating the posterior distribution of prompt embedding $p(\vz| \tilde{\gD})$. Since direct computation of posterior is intractable, in our setting we resort to variational inference (VI) and approximate the posterior distribution with a tractable surrogate $q(\vz ; \boldsymbol{\lambda})$, where $\boldsymbol{\lambda}$ denotes the variational parameters. VI minimizes KL-divergence between variational distribution and true posterior distribution with respect to $\boldsymbol{\lambda}$. i.e., $\boldsymbol{\lambda}^{\star} = \argmin_{\boldsymbol{\lambda}} \mathrm{KL}\left( q(\vz ; \boldsymbol{\lambda}) \| p(\vz| \tilde{\gD})  \right) $. This is equivalent to maximizing the evidence lower bound (ELBO), i.e., 
\begin{align} 
\boldsymbol{\lambda}^{\star}& \nonumber \\
=&\argmax_{\boldsymbol{\lambda}} \; \mathbb{E}_{q(\vz ; \boldsymbol{\lambda})}[\log p( \tilde{\gD}|\vz) ]   \\&- \mathrm{KL}\left( q(\vz ; \boldsymbol{\lambda}) \| p(\vz) \right) \nonumber \\
=& \argmax_{\boldsymbol{\lambda}} \; \sum_{i=1}^{N} \mathbb{E}_{q(\vz ; \boldsymbol{\lambda})}[\text{log Cat}(\tilde{y}_i  \vert \sigma(h_\theta( \tilde{\vx}_i; \mathbf{P}))]  \nonumber\\
&- \mathrm{KL}\left( q(\vz ; \boldsymbol{\lambda}) \| p(\vz) \right),
\label{eq:ELBO}
\end{align}
where $\mathbf{P} = \mathbf{A}\vz+\mathbf{P_0}$, and $p(\vz)$ denotes the prior distribution, which is assumed to be a normal distribution with zero mean and diagonal covariance matrix, i.e., $\gN(0,\sigma\cdot\boldsymbol{I})$.
Optimizing the ELBO objective requires taking derivative w.r.t $\boldsymbol{\lambda}$ as well as computing the gradient of log likelihood w.r.t $\vz$, i.e., $\nabla_{\vz} \mathbb{E}_{q(\vz ; \boldsymbol{\lambda})}[\text{log Cat}(\tilde{y}_i  \vert \sigma(h_\theta( \tilde{\vx}_i; \mathbf{A}\vz+\mathbf{P_0}))]$, which causes standard variational inference algorithms to be infeasible in the \textit{gradient-free} setting.

Instead of back-propagation, we propose a gradient-free variational inference algorithm leveraging the derivative-free optimizer CMA-ES. Specifically, we consider the variational distribution as a multivariate normal distribution $q(\vz ; \boldsymbol{\lambda})= \gN(\boldsymbol{\mu},\boldsymbol{\Sigma})$, where we assume the covariance matrix is diagonal, i.e.,  $\boldsymbol{\Sigma}=\operatorname{diag}(\boldsymbol{\alpha}) \in \sR^{d \times d}$. The variational parameter, as the target for optimization, is the mean and diagonal elements of the covariance matrix, i.e., $\boldsymbol{\lambda} = (\boldsymbol{\mu},\boldsymbol{\alpha}) \in \sR^{2d}$. At each iteration of the optimization, the CMA-ES outputs a collection of candidate solutions $\{\boldsymbol{\lambda}_j\}_{j=1}^m=\{ (\boldsymbol{\mu}_j,\boldsymbol{\alpha}_j) \}_{j=1}^m$.
%by sampling from another multivariate Normal distribution $\gN(m_t,\sigma^2_t C_t)$, where $m_t \in \sR^{d}$ is the current mean solution to the optimization, $\sigma_t >0$ is the current step size to update the solution, and $C_t \in \sR^{d\times d}$ is the current covariance matrix controls the shape of normal distribution. 
For each candidate variational parameter $\boldsymbol{\lambda}_j$, we evaluate the corresponding ELBO loss using the variational distribution $q(\vz ; \boldsymbol{\lambda}_j) = \gN(\boldsymbol{\mu}_j, \operatorname{diag}(\boldsymbol{\alpha}_j))$, where the expectations in Equation~\ref{eq:ELBO} is approximated by Monte-Carlo samples obtained from the variational distribution. Finally, the CMA-ES optimizer takes the current collection of variational parameter $\{\boldsymbol{\lambda}_j\}_{j=1}^m$ and their corresponding ELBO loss to conduct the next iteration of optimization. The schematic of the process is shown in Figure~\ref{fig:figure}, and the overall algorithm is summarized as Algorithm~\ref{alg:variational} in Appendix~\ref{app:algo}.

After we obtain the optimal variational parameter
$\boldsymbol{\lambda}^{\star}$ that maximizes the ELBO loss, the predictive label distribution can be estimated by taking Monte Carlo samples from the optimal variational distribution, i.e.,  $q(\vz ; \boldsymbol{\lambda}^{\star}) = \gN(\boldsymbol{\mu}^{\star},\boldsymbol{\Sigma}^{\star})$.

\subsection{SBI-based Algorithm for Likelihood-free Prompt Tuning} \label{sec:SBI}
Now, we describe our proposed approach for the \textit{gradient-free} and \textit{likelihood-free} case. For this problem, the most naive algorithm applicable is rejection approximation Bayesian computation (ABC)~\citep{pritchard1999population} that repeatedly samples from a prior distribution $\vz \sim p(\vz)$ and obtains the corresponding simulated observation $\hat{\mathbf{Y}}$. The algorithm only accepts the sampled prompt embedding if the simulated observation is sufficiently close to the ground truth observation $\tilde{\mathbf{Y}}$ based on a distance function $\rho$ and tolerance $\epsilon$, i.e., $\rho(\hat{\mathbf{Y}}, \tilde{\mathbf{Y}})< \epsilon$. The collection of accepted samples can be used to approximate the posterior distribution. However, rejection ABC typically suffers from poor computational efficiency, especially when $\epsilon$ is small and the dimensionality of observations is large. In preliminary experiments, we found rejection ABC to not be effective for our purposes. Instead, in this work, we adapt a more advanced technique --- sequential Monte Carlo approximate Bayesian computation (ABC-SMC) algorithm~\citep{mckinley2009inference} to enable efficient prompt posterior inference. The core idea of ABC-SMC is to use a sequential tolerance schedule, i.e., $\epsilon_1>\epsilon_2>,\ldots, >\epsilon_T$  to construct a sequence of intermediate distributions, which gradually converges to the true posterior distribution. 

First, we draw prompt embedding samples from the prior $p(\vz)=\gN(0,\sigma\cdot\boldsymbol{I})$ and pass them into PLM $f$ to receive the corresponding token label prediction $\hat{\mathbf{Y}}$ for a batch of text data $\tilde{\mathbf{X}}$. Then, we accept $S$ samples $\{\vz_s^{(1)}\}_{s=1}^S$ that satisfy the condition $\rho(\hat{\mathbf{Y}}, \tilde{\mathbf{Y}})< \epsilon_1$. We use accuracy as the distance function $\rho$. In the next iteration, we resample embeddings from $\{\vz_s^{(t-1)}\}_{s=1}^S$ with probability proportional to weights $w^{(t-1)}$, and perturb the sampled embeddings via a perturbation kernel to obtain a new sample, i.e., $\vz^{(t)} \sim \gN(\vz^{(t-1)}, \boldsymbol{\Sigma}^{(t-1)})$. Again, we propagate these sampled embeddings through the PLM $f$ and accept the newly proposed embeddings, $\{\vz_s^{(t)}\}_{s=1}^S$, if $\rho(\hat{\mathbf{Y}}, \mathbf{Y})< \epsilon_t$, where the tolerance $\epsilon_t$ is decayed by one step per iteration, i.e., $\epsilon_{t+1} = \epsilon_{t}-\frac 1 N$, where $N$ is the total number of training data.
Finally, the weights $w^{(t)}$ and the variance of the perturbation kernel are updated after each iteration (details are elaborated in  Appendix~\ref{app:ABC-SMC}). Empirically, we find that simply using uniform weights leads to better performance (more discussion in Section \ref{exp:dis}). These steps are repeated for $T$ iterations until the tolerance $\epsilon_T$ is sufficiently small. The schematic is in Figure~\ref{fig:figure} and the overall algorithm is summarized as Algorithm~\ref{alg:variational} in Appendix~\ref{app:algo}.

The final collection of prompt samples $\{\vz_s^{(T)}\}_{s=1}^S$ form an approximation to the posterior $p(\vz\vert\tilde{\gD})$ and we use Equation~\ref{eq:pred2} to derive the approximate predictive distribution.
%can be used as the samples drawn from the approximated posterior distribution. Since the model likelihood is not available, we use the empirical distribution to approximate the predictive label distribution, i.e., $p(\tilde{y}=c|\tilde{\vx};  \tilde{\gD}) \approx \frac{1}{S} \sum_{s=1}^{S} \mathds{1}\{\hat{y}_s=c \}$.

%% file: Sections/Exp.tex
\section{Experiment Results}
\label{sec:exp}
In this section, we demonstrate the solid empirical performance of our proposed methods. We begin with introducing the uncertainty quantification applications and describe the experiment settings. Then, we present our main results in terms of prediction performance and UQ quality. Finally, we provide an ablation study and relevant perspectives of comparison. Detailed results and implementation steps are provided in Appendix~\ref{app:results}.
\begin{table*}[!t]
  \begin{center}
  \tiny
  \caption{\textbf{Prediction Performance} (Test acc $\boldsymbol{\uparrow}$), *indicates results taken from BBT~\citep{sun2022black}}
  \vspace{-1.5em}
  \begin{tabular}{llllllllll}
    \\
    \toprule
    \textbf{Settings} & \textbf{Methods} & \textbf{SST-2} & \textbf{Yelp P.}& \textbf{AG’s News} & \textbf{DBPedia}& \textbf{MRPC} &\textbf{SNLI} &\textbf{RTE} &\textbf{Avg}\\
    \hline
    Gradient-based & Prompt Tuning* &68.23$\pm$3.78 & 61.02$\pm$6.65 & 84.81$\pm$0.66 & 87.75$\pm$1.48 & 51.61$\pm$8.67 & 36.13$\pm$1.51 & 54.69$\pm$3.79 &  63.46  \\
    & P-Tuning v2* &  64.33$\pm$3.05 & 92.63$\pm$1.39 & 83.46$\pm$1.01 & 97.05$\pm$0.41 & 68.14$\pm$3.89 & 36.89$\pm$0.79 & 50.78$\pm$2.28 & 70.47  \\
    &Model Tuning* & 85.39$\pm$2.84& 91.82$\pm$0.79& 86.36$\pm$1.85 &97.98$\pm$0.14 & 77.35$\pm$5.70 & 54.64$\pm$5.29 & 58.60$\pm$6.21 &78.88\\
    \hline
    Gradient-free & Manual Prompt*  &  79.82 & 89.65 & 76.96 & 41.33 & 67.40 & 31.11 & 51.62 & 62.56  \\
              & In-Context Learning* & 79.79$\pm$3.06 & 85.38$\pm$3.92 & 62.21$\pm$13.46 & 34.83$\pm$7.59 & 45.81$\pm$6.67 & 47.11$\pm$0.63 & 60.36$\pm$1.56 & 59.36  \\
              & Feature-MLP* & 64.80$\pm$1.78 & 79.20$\pm$2.26 & 70.77$\pm$0.67 & 87.78$\pm$0.61 & 68.40$\pm$0.86 & 42.01$\pm$0.33 & 53.43$\pm$1.57 & 66.63 \\
              & Feature-BiLSTM* &   65.95$\pm$0.99 & 74.68$\pm$0.10 
 &77.28$\pm$2.83 & 90.37$\pm$3.10 & 71.55$\pm$7.10 & 46.02$\pm$0.38 & 52.17$\pm$0.25 & 68.29  \\
   
    & BBT & 86.93$\pm$0.25 & 91.61$\pm$0.29 & 83.22$\pm$0.42 & 76.94$\pm$1.22 & 75.95$\pm$2.30 & 45.38$\pm$0.02 & 50.54$\pm$0.36 & 72.94  \\
              & \textbf{Ours(ELBO)} & 86.81$\pm$0.47 & 92.07$\pm$0.17 & 83.96$\pm$0.22 & 73.25$\pm$2.35 & 76.35$\pm$0.94 & 46.78$\pm$2.92 & 50.78$\pm$1.39 & 72.86 \\
              & \textbf{Ours(Ensembles)}     & 88.61$\pm$0.78 & 92.35$\pm$0.16 & 84.62$\pm$0.20 & 80.12$\pm$1.06 & 76.77$\pm$1.13 & 47.95$\pm$2.76 & 50.34$\pm$3.40 & \textbf{74.39}\\
    \hline \hline
    \makecell{Gradient-free} \&  & \textbf{Ours(SNPE)} & 84.37$\pm$0.29 & 90.38$\pm$0.07 & 80.50$\pm$0.10 & 33.11$\pm$0.48 & 81.02$\pm$0.06 & 39.60$\pm$0.49 & 53.07$\pm$0.82 & 66.01\\
    \makecell{Likelihood-free} & \textbf{Ours(ABC-SMC)} & 86.51$\pm$0.55 & 90.32$\pm$0.03 & 81.43$\pm$0.41 & 57.41$\pm$0.90 & 80.78$\pm$0.07 & 40.81$\pm$0.24 & 53.37$\pm$0.30 & \textbf{70.09}\\

    \toprule
  \end{tabular}
  \label{table:test-acc}
  \end{center}
  \vspace{-1em}
\end{table*}

\begin{table*}[!t]
  \begin{center}
  \tiny
  \caption{\textbf{Calibration Performance} (ECE score $\boldsymbol{\downarrow}$)}
  \vspace{-1.5em}
  \begin{tabular}{llllllllll}
    \\
    \toprule
    \textbf{Settings} & \textbf{Methods} & \textbf{SST-2} & \textbf{Yelp P.}& \textbf{AG’s News} & \textbf{DBPedia}& \textbf{MRPC} &\textbf{SNLI} &\textbf{RTE} &\textbf{Avg}\\
    \hline
    Gradient-free & BBT & 0.056$\pm$0.014 & 0.032$\pm$0.000 & 0.049$\pm$0.007 & 0.056$\pm$0.032 & 0.115$\pm$0.018 & 0.040$\pm$0.008 & 0.170$\pm$0.069 & 0.074  \\
              & \textbf{Ours(ELBO)} & 	0.056$\pm$0.007 & 0.025$\pm$0.004 & 0.065$\pm$0.001 & 0.045$\pm$0.028 & 0.058$\pm$0.004 & 0.035$\pm$0.007 & 0.113$\pm$0.030 & \textbf{0.057}\\
              & \textbf{Ours(Ensembles)}     & 0.058$\pm$0.001 & 0.017$\pm$0.001 & 0.064$\pm$0.009 & 0.085$\pm$0.005 & 0.073$\pm$0.007 & 0.039$\pm$0.004 & 0.134$\pm$0.033 & 0.067\\
    \hline \hline
    \makecell{Gradient-free \& }   & \textbf{Ours(SNPE)} & 0.104$\pm$0.005 & 0.082$\pm$0.000 & 0.100$\pm$0.010 & 0.549$\pm$0.004 & 0.314$\pm$0.001 & 0.185$\pm$0.011 & 0.466$\pm$0.002 & 0.257\\
    \makecell{Likelihood-free} & \textbf{Ours(ABC-SMC)} & 0.106$\pm$0.009 & 0.084$\pm$0.001 & 0.108$\pm$0.001 & 0.278$\pm$0.026 & 0.309$\pm$0.009 & 0.178$\pm$0.002 & 0.458$\pm$0.004 & \textbf{0.217}\\
    \toprule
  \end{tabular}
  \label{table:calibration}
  \end{center}
   \vspace{-1em}
\end{table*}

\begin{table*}[!t]
  \begin{center}
  \tiny
  \caption{\textbf{Selective Classification} (AURRRC score $\boldsymbol{\downarrow}$)}
  \vspace{-1.5em}
  \begin{tabular}{llllllllll}
    \\
    \toprule
    \textbf{Settings} & \textbf{Methods} & \textbf{SST-2} & \textbf{Yelp P.}& \textbf{AG’s News} & \textbf{DBPedia}& \textbf{MRPC} &\textbf{SNLI} &\textbf{RTE} &\textbf{Avg}\\
    \hline
    & \textbf{Lower-bound} & 0.030 & 0.009 &  0.035 & 0.070 & 0.251 & 0.427 & 0.255 & 0.154  \\
    \hline
    Gradient-free& BBT(Entropy) & 0.063$\pm$0.009 & 0.029$\pm$0.001 & 0.082$\pm$0.004 & 0.095$\pm$0.009 & 0.349$\pm$0.002 & 0.519$\pm$0.032 & 0.523$\pm$0.004 & 0.237  \\
    & BBT(MaxP) & 0.063$\pm$0.009 & 0.029$\pm$0.001 & 0.077$\pm$0.004 & 0.091$\pm$0.009 & 0.349$\pm$0.002 & 0.513$\pm$0.031 & 0.523$\pm$0.004 & 0.235  \\
              & \textbf{ELBO(Entropy)} & 0.053$\pm$0.004 & 0.026$\pm$0.001 & 0.079$\pm$0.001 & 0.123$\pm$0.006 & 0.336$\pm$0.009 & 0.481$\pm$0.065 & 0.508$\pm$0.012 & 0.229\\
              & \textbf{ELBO(MaxP)} & 0.053$\pm$0.004 & 0.026$\pm$0.001 & 0.074$\pm$0.002 & 0.117$\pm$0.005 & 0.336$\pm$0.009 & 0.478$\pm$0.062 & 0.508$\pm$0.012 & 0.227\\
              & \textbf{Ensembles(Entropy)}     & 0.046$\pm$0.006 & 0.023$\pm$0.001 & 0.074$\pm$0.002 & 0.084$\pm$0.004 & 0.324$\pm$0.011 & 0.472$\pm$0.048 & 0.513$\pm$0.048 & 0.219\\
              & \textbf{Ensembles(MaxP)}     & 0.046$\pm$0.006 & 0.023$\pm$0.001 & 0.068$\pm$0.002 & 0.076$\pm$0.004 & 0.324$\pm$0.011 & 0.469$\pm$0.047 & 0.513$\pm$0.048 & \textbf{0.217}\\
    \hline \hline
    \makecell{Gradient-free \& }  & \textbf{SNPE(Entropy)} & 0.065$\pm$0.003 & 0.073$\pm$0.001 & 0.116$\pm$0.005 & 0.551$\pm$0.001 & 0.319$\pm$0.003 & 0.580$\pm$0.009 & 0.466$\pm$0.003 & 0.310\\
    \makecell{Likelihood-free} & \textbf{SNPE(MaxP)} & 0.065$\pm$0.003 & 0.073$\pm$0.001 & 0.116$\pm$0.005 & 0.552$\pm$0.002 & 0.319$\pm$0.003 & 0.591$\pm$0.009 & 0.466$\pm$0.003 & 0.312\\
    & \textbf{ABC-SMC(Entropy)} & 0.061$\pm$0.006 & 0.075$\pm$0.002 & 0.110$\pm$0.004 & 0.285$\pm$0.015& 0.325$\pm$0.006  & 0.571$\pm$0.000 & 0.468$\pm$0.014 & \textbf{0.271}\\
    & \textbf{ABC-SMC(MaxP)} & 0.061$\pm$0.006 & 0.075$\pm$0.002 & 0.110$\pm$0.004 & 0.288$\pm$0.014& 0.325$\pm$0.006  & 0.579$\pm$0.000 & 0.468$\pm$0.014 & 0.272\\
    \toprule
  \end{tabular}
  \label{table:selective}
  \end{center}
   \vspace{-1em}
\end{table*}

\begin{table*}[h]
  \begin{center}
  \tiny
  \caption{\textbf{Far OOD Detection} (AURRRC score $\boldsymbol{\downarrow}$)}
  \vspace{-1.5em}
  \begin{tabular}{llllllllll}
    \\
    \toprule
    \textbf{Settings} & \textbf{Methods} & \makecell[l]{ID:\textbf{SST-2}\\ OOD:\textbf{RTE}} & \makecell[l]{ID:\textbf{Yelp P.}\\ OOD:\textbf{RTE}} &\makecell[l]{ID:\textbf{MRPC}\\ 
    OOD:\textbf{RTE}}&  \makecell[l]{ID:\textbf{DBPedia}\\ OOD:\textbf{AG’s News}}
    &\makecell[l]{ID:\textbf{SNLI}\\ OOD:\textbf{MRPC}}& \makecell[l]{ID:\textbf{RTE}\\ OOD:\textbf{MRPC}} &\textbf{Avg}\\
    \hline
    & \textbf{Lower-bound} & 0.072 & 0.001 & 0.162 & 0.010 & 0.004 & 0.357 & 0.101   \\
    \hline
    Gradient-free& BBT(entropy) & 0.124$\pm$0.015 & 0.002$\pm$0.000 & 0.404$\pm$0.006 & 0.058$\pm$0.018 & 0.100$\pm$0.002 & 0.639$\pm$0.024 &0.221 \\
             & BBT(MaxP) & 0.124$\pm$0.015 & 0.002$\pm$0.000 & 0.404$\pm$0.006& 0.059$\pm$0.014 & 0.098$\pm$0.002 & 0.639$\pm$0.024 &0.221 \\
              & \textbf{Ours(ELBO)(Entropy)} & 0.112$\pm$0.010 & 0.001$\pm$0.000 & 0.320$\pm$0.014 & 0.051$\pm$0.001 & 0.109$\pm$0.002 & 0.635$\pm$0.003 &0.205 \\
              & \textbf{Ours(ELBO)(MaxP)} & 0.112$\pm$0.010 & 0.001$\pm$0.000 & 0.320$\pm$0.014 & 0.056$\pm$0.001 & 0.107$\pm$0.002 & 0.635$\pm$0.003 &0.205 \\
              & \textbf{Ours(Ensembles)(Entropy)}     & 0.097$\pm$0.008 & 0.001$\pm$0.000 & 0.350$\pm$0.038 & 0.057$\pm$0.003 & 0.110$\pm$0.001  & 0.606$\pm$0.047& 0.204 \\
              & \textbf{Ours(Ensembles)(MaxP)}   & 0.097$\pm$0.008 & 0.001$\pm$0.000 & 0.350$\pm$0.038 & 0.058$\pm$0.002 & 0.108$\pm$0.001  & 0.606$\pm$0.047& \textbf{0.203} \\
    \hline \hline
    \makecell{Gradient-free \&}   & \textbf{Ours(SNPE)(Entropy)} & 0.140$\pm$0.001 & 0.005$\pm$0.000 &  0.402$\pm$0.005 & 0.082$\pm$0.003 &  0.093$\pm$0.001 & 0.592$\pm$0.008 &0.219\\
    \makecell{Likelihood-free} & \textbf{Ours(SNPE)(MaxP)} & 0.140$\pm$0.001 & 0.005$\pm$0.000 & 0.402$\pm$0.005 & 0.081$\pm$0.003 &  0.091$\pm$0.002 & 0.592$\pm$0.008 &0.219\\
    & \textbf{Ours(ABC-SMC)(Entropy)} & 0.126$\pm$0.009 & 0.005$\pm$0.001 & 0.396$\pm$0.001& 0.097$\pm$0.021 & 0.092$\pm$0.000 & 0.596$\pm$0.009 &0.219\\
    & \textbf{Ours(ABC-SMC)(MaxP)} & 0.126$\pm$0.009  & 0.005$\pm$0.001 & 0.396$\pm$0.001 & 0.095$\pm$0.021& 0.092$\pm$0.001 & 0.596$\pm$0.009 &\textbf{0.218}\\
    \toprule
  \end{tabular}
  \label{table:far-OOD-detection}
  \end{center}
   \vspace{-1em}
\end{table*}

\begin{table*}[h]
  \begin{center}
  \tiny
  \caption{\textbf{Near OOD Detection} (AURRRC score $\boldsymbol{\downarrow}$)}
  \vspace{-1.5em}
  \begin{tabular}{llllllll}
    \\
    \toprule
    \textbf{Settings} & \textbf{Methods} & \makecell[l]{ID:\textbf{SST-2}\\ OOD:\textbf{IMDB}} & \makecell[l]{ID:\textbf{Yelp P.}\\ OOD:\textbf{IMDB}} &\makecell[l]{ID:\textbf{SNLI}\\ OOD:\textbf{MNLI}}& \makecell[l]{ID:\textbf{RTE}\\ OOD:\textbf{MNLI}} &\textbf{Avg}\\
    \hline
    & \textbf{Lower-bound} &  0.960 &  0.147 & 0.259 & 0.950 & 0.579    \\
    \hline
    Gradient-free & BBT(entropy) & 0.978$\pm$0.003 & 0.315$\pm$0.003 & 0.720$\pm$0.011 & 0.963$\pm$0.008 & 0.744  \\
              & BBT(confidence) & 0.978$\pm$0.003 & 0.315$\pm$0.003 & 0.705$\pm$0.011 & 0.963$\pm$0.008 & 0.740  \\
              & \textbf{Ours(ELBO)(Entropy)} & 0.976$\pm$0.002 & 0.308$\pm$0.006 & 0.692$\pm$0.039 & 0.968$\pm$0.005 & 0.736  \\
              & \textbf{Ours(ELBO)(MaxP)} & 0.976$\pm$0.002 & 0.308$\pm$0.006 & 0.678$\pm$0.044 & 0.968$\pm$0.005 & 0.733  \\
              & \textbf{Ours(Ensembles)(Entropy)}     & 0.976$\pm$0.001 & 0.297$\pm$0.003 & 0.707$\pm$0.028 &  0.962$\pm$0.002 & 0.736 \\
              & \textbf{Ours(Ensembles)(MaxP)}     & 0.976$\pm$0.001 & 0.297$\pm$0.003 & 0.692$\pm$0.032 & 0.962$\pm$0.002 & \textbf{0.732} \\
    \hline \hline
    \makecell{Gradient-free \&} & \textbf{Ours(SNPE)(Entropy)} & 0.984$\pm$0.000 & 0.365$\pm$0.001 & 0.715$\pm$0.002 & 0.951$\pm$0.000 & 0.754 \\
    \makecell{Likelihood-free}& \textbf{Ours(SNPE)(MaxP)} & 0.984$\pm$0.000 & 0.365$\pm$0.001 & 0.695$\pm$0.004 & 0.951$\pm$0.000 & 0.749 \\
    & \textbf{Ours(ABC-SMC)(Entropy)} & 0.983$\pm$0.001 & 0.365$\pm$0.001 & 0.710$\pm$0.002 & 0.952$\pm$0.000 & 0.753 \\
    & \textbf{Ours(ABC-SMC)(MaxP)} & 0.983$\pm$0.001 & 0.365$\pm$0.001 & 0.694$\pm$0.000 & 0.952$\pm$0.000 & \textbf{0.749} \\
    \toprule
  \end{tabular}
  \label{table:near-OOD-detection}
  \end{center}
   \vspace{-1em}
\end{table*}

\subsection{Settings}
\paragraph{Uncertainty Quantification Applications.}
We assess the performance of the uncertainty quantification from three perspectives: (1) \textbf{Calibration} -- the typical UQ quality metric that measures how well the model confidence aligned with the correctness of its prediction; (2) \textbf{Selective Classification} -- aims to avoid the risk of wrong predictions by abstaining the prediction for samples with high uncertainty; and (3) \textbf{OOD Detection} -- aims to identify the out-of-distribution data that is unobserved during the training stage. The OOD data can exhibit different forms of distribution shift, including the \textit{covariate shift} where the OOD data distribution is different from the training samples; and the \textit{semantic shift} where the OOD data contain unobserved class. In our experiment, we focus on two types of OOD tasks: the \textbf{Far OOD} detection task where both \textit{covariat shift} and \textit{semantic shift} happen simultaneously; the \textbf{Near OOD} detection task where the OOD data only contain \textit{covariate shift}, but have the same class label words.

\paragraph{Benchmark.}
For a comprehensive comparison with BBT~\citep{sun2022black}, we mainly employ the same text classification benchmark datasets as BBT, including sentiment analysis datasets SST-2~\citep{socher2013recursive} and Yelp polarity~\citep{zhang2015character};  topic classification datasets AG’s News~\citep{zhang2015character} and DBPedia~\citep{zhang2015character}; paraphrase
dataset MRPC~\citep{dolan2005automatically}; natural language
inference (NLI) datasets SNLI~\citep{bowman2015large} and RTE~\citep{wang2018glue}.

Both calibration and selective classification tasks are conducted using the original test samples for each benchmark dataset. For the far OOD detection task, we create the ID/OOD dataset pairs by combining two datasets belonging to two different tasks, e.g., SST-2/RTE. For the near OOD detection task, we use IMDB~\citep{maas2011learning} for the sentiment analysis task and MNLI~\citep{williams2017broad} for the NLI task.
\paragraph{Baselines.}
%We demonstrate our proposed methods' strong empirical performance in terms of prediction and uncertainty quantification performance. 
For prediction performance, besides the SOTA \textit{Gradient-free} prompt tuning approach BBT~\citep{sun2022black}, we also compare with other \textit{Gradient-free} methods: (1) The naive \textbf{Manual Prompt} that uses the hand-crafted prompt templates; (2) \textbf{In-context Learning}~\citep{brown2020language}; (3) Feature-based approaches~\citep{peters2019tune} that trains auxiliary models on top of the PLM extracted features, including \textbf{Feature-MLP} training a MLP classifier and \textbf{Feature-BiLSTM} training a LSTM model followed by a classifier. We include additional results of \textit{Gradient-based} approaches: (1) \textbf{Model Tuning} that fine-tunes the entire PLM; (2) \textbf{Prompt Tuning}~\citep{lester2021power} that only trains the continuous prompt without modifying PLM; (3) \textbf{P-Tuning v2}~\citep{liu2021p} that trains the several continuous prompts injected at different layers of PLM. For uncertainty quantification tasks, few existing prompt tuning works aim to tackle this problem, so we mainly compare with BBT to justify how we can address its limitation under the \textit{gradient-free} setting.
\paragraph{Implementation Details.}
We follow the same experiment setting as BBT. We focus on text classification as a few-shot learning problem, motivated by the fact that labeled training data can be limited in practice. Specifically, we construct few-shot training and validation data by drawing $16$ random samples for each class from the original training dataset. The prediction performance is evaluated on the original development or test set, depending on the datasets. We use the same PLM model $\operatorname{RoBERTa_{LARGE}}$ as the backbone model and keep the hyper-parameter same as BBT. Specifically, we set the prompt length as $50$, i.e., $D = 50 \times 1024$, and the subspace dimensionality as $d = 500$. The only modification is that we adapt the normal distribution~\citep{sun2022bbtv2} to generate the random projection matrix $\mathbf{A}$, instead of the uniform distribution used in BBT. For a fair comparison, we reproduce the results of BBT using the random projection generated from normal distribution. More implementation details are included in Appendix~\ref{app:hyper}.

\paragraph{Performance Metrics.}
For prediction performance, we evaluate the prediction accuracy on the testing dataset. For calibration performance, we adopt the expected calibration error (ECE)~\citep{guo2017calibration} score as the metric. For both selective classification and OOD detection tasks, we compute the area under the risk vs. rejection rate curve (AURRRC)~\citep{franc2019discriminative}. The risk is defined as the portion of wrong-predicted samples among the data chosen for prediction in selective classification and the portion of OOD samples among the data identified as in-distribution in the OOD detection task. The rejection rate is defined as the portion of data that abstained from the prediction based on specific uncertainty measurement. Note that an oracle with perfect knowledge of uncertainty measurement can achieve a minimum AURRRC score. This is obtained by assigning an uncertainty score based on the oracle knowledge, i.e., whether a test sample is wrong-predicted (OOD samples) or not. We denote such minimum AURRRC score as the \textit{lower-bound}.

Given the predictive label distribution, we utilize two uncertainty measurements, including \textit{Entropy} of the label distribution, i.e., $\gH\left(p(\tilde{y}|\tilde{\vx}; \tilde{\gD}) \right)$, and \textit{MaxP}, which is defined as $\max_c p(\tilde{y}=c|\tilde{\vx}; \tilde{\gD})$.

\subsection{Results}
We conduct extensive evaluations of our proposed methods under both the \textit{Gradient-free} setting and the \textit{Gradient-free} and \textit{Likelihood-free} setting. The results of prediction performance are shown in Table~\ref{table:test-acc}. For the uncertainty quantification performance, the calibration results are shown in Table~\ref{table:calibration}, the selective classification results are shown in Table~\ref{table:selective}, the Far OOD detection results are shown in Table~\ref{table:far-OOD-detection} and the Near  OOD detection results are shown in Table~\ref{table:near-OOD-detection}.

\paragraph{Gradient-free and Likelihood-free Setting.}
No existing work is trying to tackle the \textit{Gradient-free} and \textit{likelihood-free} prompt tuning problem. However, we still compare our proposed method with other baseline methods on different problem settings to understand how well we can achieve and the price we need to pay for such a more strict constraint. In addition, we also include the results of neural net-based approach SNPE~\citep{hermans2020likelihood, durkan2020contrastive} for solving the SBI problem.

As shown in Table~\ref{table:test-acc}, our proposed method ABC-SMC can achieve competitive prediction performance as SOTA approach BBT and even outperform the other \textit{Gradient-free} baselines without the requirement of the model likelihood. We also observe that ABC-SMC performs better than SNPE. The possible explanation is that the density estimation model adopted by SNPE usually requires a large number of simulated samples to achieve good performance, which is hindered by the slow inference speed of large PLM.

For the uncertainty quantification tasks, ABC-SMC underperforms on calibration and selective classification tasks but can still achieve comparable performance on the two OOD detection tasks. The performance gap can possibly be mitigated if we collect more samples (by increasing $K$) for a more accurate estimation of the empirical label distribution, but the computational cost is the price we need to pay for the \textit{likelihood-free} constraint.

\paragraph{Gradient-free Setting.}
By relaxing the \textit{likelihood-free} constraint, it is observed that our proposed methods, both Gradient-free Variational Inference (denoted as ELBO) and Ensembles algorithms, achieve comparable or even better prediction performance than BBT and other \textit{gradient-free} baselines, while outperforming BBT in terms of uncertainty quantification across all the tasks. Such empirical observation justifies the effectiveness of leveraging Bayesian and Ensemble techniques to enable more reliable \textit{gradient-free} prompt tuning without sacrificing the prediction performance.

\subsection{Discussions} \label{exp:dis}
In this section, we further investigate our proposed methods by exploring the use of alternate models and the effect of using uniform weights in the SMC-ABC algorithm. 
\paragraph{Performance on other backbone models}
To demonstrate that our proposed methods generalize well on other PLM backbone models, we evaluate them on $\operatorname{BERT_{LARGE}}$ under the both \textit{Gradient-free} setting and \textit{Gradient-free} and \textit{likelihood-free} setting. The results are presented in Appendix~\ref{app:backbone}. Note that our proposed methods consistently outperform BBT in terms of both prediction and uncertainty quantification performance under the \textit{Gradient-free} setting while achieving competitive performance with a small gap under the \textit{Gradient-free} and \textit{likelihood-free} setting.

\paragraph{Ablation study of weights in ABC-SMC}
In practice, we observe that the ABC-SMC algorithm suffers from weight degeneracy, with weights for certain particles approaching one and effectively causing the posterior to be approximated by a single particle. Although, this issue can be mitigated by designing better proposals, we found that the heuristic of using uniform weights instead of updating the weights at each iteration of the algorithm to be far more effective. To demonstrate the efficacy, we conduct an ablation study about the sampling weights, and the results are shown in Appendix~\ref{app:ablation-weights}. We find that with uniform weights ABC-SMC provides both improves prediction and uncertainty quantification for our application.

%% file: Sections/conclusion.tex
\section{Concluding Remarks}
% In this work, we explore \textit{gradient-free} prompt tuning along two under-explored angles: quantifying uncertainty in soft prompts; and tackling a more strict \textit{likelihood-free} setting from the SBI perspective. Our developed methods demonstrate encouraging empirical performance across multiple tasks. 

% %While the key idea of this work is to learn the posterior distribution of prompts, evaluating the quality of the approximated posterior from the inference algorithms is a challenging problem. This may result in overestimating or underestimating the uncertainty. Meanwhile, it is observed that a performance gap remains between the \textit{likelihood-free} setting and \textit{likelihood} setting. Thus, 
% Investigating more modern neural SBI methods and designing more robust methods for learning prompt posteriors are interesting directions for future research. Other perspectives on \textit{gradient-free} prompt tuning, such as learning natural language-like interpretable prompts are also worthy of exploration. 
%
In this work, we explore \textit{gradient-free} prompt tuning along two under-explored angles: quantifying uncertainty in soft prompts; and tackling a more strict \textit{likelihood-free} setting from the SBI perspective. Our developed methods demonstrate encouraging empirical performance across multiple tasks. 

Investigating more modern neural SBI methods and designing more robust methods for learning prompt posteriors are exciting directions for future research. Other perspectives on \textit{gradient-free} prompt tuning, such as learning natural language-like interpretable prompts, are also worthy of exploration.

%% file: Sections/Appendix.tex
\section{Omitted Algorithms} \label{app:algo}
The overall algorithm of Gradient-free Variational Inference and ABC-SMC are shown in Algorithm~\ref{alg:variational} and Algorithm \ref{alg:ABC-SMC}, respectively.
\begin{algorithm}[h]
\small
\SetAlgoLined
    \textbf{input}{ Training data set $\{(\tilde{\vx}_i, \tilde{y}_i)\}_{i=1}^{N}$; CMA-ES optimizer $\operatorname{ES}$; Prior distribution $p(\vz)$; Number of candidate solutions $m$; Total iteration $T$. }

Initialize the initial collection of variational parameter, i.e., $\{\boldsymbol{\lambda}^{(0)}_j\}_{j=1}^m=\{ (\boldsymbol{\mu}^{(0)}_j,\boldsymbol{\alpha}^{(0)}_j) \}_{j=1}^m$. \\
\For{$t = 1,2, \ldots, T$}
{   
    \For{$j = 1,2, \ldots, m$}
    {
        Generate $S$ prompt embedding samples from the variational distribution, i.e.,  $$\{\vz^{(t-1)}_s\}_{s=1}^S \sim \gN\left(\boldsymbol{\mu}_j^{(t-1)},\operatorname{diag}(\boldsymbol{\alpha}_j^{(t-1)})\right)$$
    
        Evaluate the ELBO loss of $j$-th variational distribution i.e., \begin{align}
        &\gL_j^{(t-1)} \nonumber\\
        &= \sum_{i=1}^{N}\sum_{s=1}^{S}\text{log Cat}(\tilde{y}_i  \vert \sigma(h_\theta( \tilde{\vx}_i; \mathbf{A}\vz^{(t-1)}_s+\mathbf{P_0}))\nonumber\\
        &- \mathrm{KL}\left( q(\vz ; \boldsymbol{\lambda}_j^{(t-1)}) \| p(\vz) \right) \nonumber
        \end{align}
    }
    Request a new collection of variational parameter solutions, i.e., $$\{\boldsymbol{\lambda}^{(t)}_j\}_{j=1}^m \leftarrow \operatorname{ES}\left(\{\boldsymbol{\lambda}^{(t-1)}_j\}_{j=1}^m; \{\gL_j^{(t-1)}\}_{j=1}^m\right)$$
}
    \textbf{output}{ Optimized collection of prompt embedding samples $\{\vz^{(T)}_s\}_{s=1}^S$ corresponding to max ELBO loss.}

 \caption{Gradient-free Variational Inference}\label{alg:variational}
\end{algorithm}

\begin{algorithm}[h]
\small
\SetAlgoLined
    \textbf{input}{ PLM $f$; The fixed random projection matrix $\mathbf{A}$ and intial prompt $\mathbf{P_0}$; Training data set $(\tilde{\mathbf{X}},\tilde{\mathbf{Y}}) = \{(\tilde{\vx}_i, \tilde{y}_i)\}_{i=1}^{N}$; Prior distribution $p(\vz)$; Initial tolerance $\epsilon_1$; Distance measure function $\rho(\cdot)$; Number of samples $S$; Total iteration $T$. }

\For{$t = 1,2, \ldots, T$}
{   
    \eIf{$t==1$}
    {
        \For{$s = 1,2, \ldots, S$}
        {
            \Do{$\rho(\hat{\mathbf{Y}}, \tilde{\mathbf{Y}})> \epsilon_1$}
            {
            Generate prompt embedding samples from the prior distribution, i.e.,  $\vz_s^{(1)} \sim p(\vz)$;

            Obtain the corresponding prediction result $\hat{\mathbf{Y}} = f(\mathbf{A}\vz_s^{(1)}+\mathbf{P_0};\tilde{\mathbf{X}})$.
            }
            Initialize the sampling probability weights $w_s^{(1)}=\frac{1}{S}$.
        }
        Decay the tolerance, i.e.,  $\epsilon_{t+1} = \epsilon_{t}-\frac 1 N$; Initialize the perturbation kernel variance $\boldsymbol{\Sigma}^{(1)}$.
    }
    {
        \For{$s = 1,2, \ldots, S$}
        {
            \Do{$\rho(\hat{\mathbf{Y}}, \tilde{\mathbf{Y}})> \epsilon_t$}
            {
                Draw a random sample $\vz_s^{(t-1)}$ from $\{\vz^{(t-1)}_s\}_{s=1}^S$ with probability $w_s^{(t-1)}$;

                Generate a new sample $\vz_s^{(t)} \sim \gN(\vz_s^{(t-1)}, \boldsymbol{\Sigma}^{(t-1)})$;

                Obtain the corresponding prediction result $\hat{\mathbf{Y}} = f(\mathbf{A}\vz_s^{(t)}+\mathbf{P_0};\tilde{X})$.
            }
            Update the sampling probability weights $w^{(t)}$ (see Appendix \ref{app:ABC-SMC}).
        }
        Decay the tolerance, i.e.,  $\epsilon_{t+1} = \epsilon_{t}-\frac 1 N$; Update the perturbation kernel variance $\boldsymbol{\Sigma}^{(t)}$ (see Appendix \ref{app:ABC-SMC}). 
    }
}
    \textbf{output}{ Optimized collection of prompt embedding samples $\{\vz^{(T)}_s\}_{s=1}^S$; Final sampling weights $w^{(T)}$. }

 \caption{ABC-SMC}\label{alg:ABC-SMC}
\end{algorithm}

\section{Implementation details of ABC-SMC } \label{app:ABC-SMC}
\paragraph{Updating of $w^{(t)}$}
In the ABC-SMC algorithm, the sampling weights are initialized as uniform distribution at the first iteration $t=1$ as all the samples are sampled from the prior distribution $p(\vz)$. In the later iterations, the new samples are drawing from a mixture proposal distribution consisted the previous samples and the perturbation kernel, i.e., $\sum_{s=1}^{S}  w_s^{(t-1)} \cdot \gN(\vz_s^{(t-1)}, \boldsymbol{\Sigma}^{(t-1)})$. The weights are updated in an importance sampling manner as the ratio between the prior probability and the proposal probability, i.e.,
$$
w_s^{(t)} = \frac{p(\vz_s)}{\sum_{s=1}^{S}  w_s^{(t-1)} \cdot \gN(\vz_s^{(t-1)}, \boldsymbol{\Sigma}^{(t-1)})} 
$$
\paragraph{Updating of $\boldsymbol{\Sigma}^{(t)}$}
The covariance $\boldsymbol{\Sigma}^{(t)}$ in the perturbation kernel is a diagonal covariance matrix $\operatorname{diag}(\boldsymbol{\alpha}^{(t)})$, where the diagonal elements $\boldsymbol{\alpha}^{(t)}$ are updated using the weighted empirical variance of previous collection of samples, i.e. 
$$
\boldsymbol{\alpha}^{(t)} = \sum_{s=1}^{S}  w_s^{(t-1)} \cdot (\vz_s^{(t-1)}-\bar{\vz}^{(t-1)})^2
$$
Where $\bar{\vz}^{(t-1)}=\sum_{s=1}^{S}  w_s^{(t-1)} \cdot \vz_s^{(t-1)}$ is the mean.

\clearpage
\section{Implementation Details} \label{app:hyper}
All of our experiment results are reported with means and standard deviations over three trials, each with a different random seed. The experiments are implemented in PyTorch, and each run of our proposed methods requires less than 24h of training computation time (on a single NVIDIA Tesla V100 GPU). Our proposed algorithms generate a collection of $S$ prompt samples to estimate the predictive label distribution. We set $S=10, 100, 100$ for Ensembles, Gradient-free Variational Inference, and ABC-SMC, respectively. The total budget for the derivative-free optimizer CMA-ES is set to be 300 with a population size of 20. We use the same prior distribution $p(\vz)$ for all algorithms, which is assumed to be a normal distribution with zero mean and diagonal covariance matrix, i.e., $\gN(0,\sigma\cdot\boldsymbol{I})$. $\sigma$ controls how concentrated the prior distribution is, and we use $\sigma=50$ in our experiments. In ABC-SMC, the distance measure function $\rho$ is defined as the prediction error rate, i.e., the portion of wrongly predicted data among the whole data batch. The initial tolerance $\epsilon_1$ in ABC-SMC is initialized as the prediction error rate of an arbitrary prompt sample drawing from the prior distribution. The tolerance is decayed by one step per iteration, i.e., $\epsilon_{t+1} = \epsilon_{t}-\frac 1 N$, where $N$ is the total number of training data.

\section{Additional Experiment Results}  \label{app:results}

\subsection{Performance on other backbone PLM} \label{app:backbone}
We evaluate the performance of our proposed methods on SST-2 and SNLI tasks using $\operatorname{BERT_{LARGE}}$ as the backbone model. We keep the hyper-parameter settings the same as the original experiments. The results are shown in Table~\ref{table:test-acc-backbone}, \ref{table:calibration-backbone}, \ref{table:selective-backbone}, and \ref{table:far-OOD-detection-backbone}.
\begin{table}[h]
  \begin{center}
  \tiny
  \caption{\textbf{Test Performance (test acc $\uparrow$)}}
  \vspace{-1.5em}
  \begin{tabular}{llll}
    \\
    \toprule
    \textbf{Settings} & \textbf{Methods} & \textbf{SST-2} &\textbf{SNLI}\\
    \hline
    Gradient-free& BBT & 74.77$\pm$3.21 & 41.07$\pm$2.97   \\
              & \textbf{Ours(ELBO)} & 75.38$\pm$1.74 & 41.20$\pm$0.39 \\
              & \textbf{Ours(Ensembles)}     & 80.05$\pm$1.79& 42.64$\pm$1.96 \\
    \hline \hline
    \makecell{Gradient-free \& \\ Likelihood-free} & \textbf{Ours(ABC-SMC)} & 66.40$\pm$0.46 & 39.00$\pm$0.22\\
    \toprule
  \end{tabular}
  \label{table:test-acc-backbone}
  \end{center}
 
\end{table}

\begin{table}[h]
  \begin{center}
  \tiny
   \caption{\textbf{Calibration Performance} (ECE score $\boldsymbol{\downarrow}$)}
   \vspace{-1.5em}
  \begin{tabular}{llll}
    \\
    \toprule
    \textbf{Settings} & \textbf{Methods} & \textbf{SST-2} &\textbf{SNLI}\\
    \hline
    Gradient-free& BBT & 0.081$\pm$0.051 & 0.086$\pm$0.039   \\
              & \textbf{Ours(ELBO)} & 0.046$\pm$0.006 & 0.073$\pm$0.009 \\
              & \textbf{Ours(Ensembles)}     & 0.045$\pm$0.007 & 0.068$\pm$0.024 \\
    \hline \hline
    \makecell{Gradient-free \& \\ Likelihood-free} & \textbf{Ours(ABC-SMC)} & 0.328$\pm$0.003 & 0.584$\pm$0.002\\
    \toprule
  \end{tabular}
  \label{table:calibration-backbone}
  \end{center}
  
\end{table}

\begin{table}[!t]
  \begin{center}
  \tiny
  \caption{\textbf{Selective Classification} (AURRRC score $\boldsymbol{\downarrow}$)}
  \vspace{-1.5em}
  \begin{tabular}{llllllllll}
    \\
    \toprule
    \textbf{Settings} & \textbf{Methods} & \textbf{SST-2} &\textbf{SNLI} \\
    \hline
    Gradient-free& BBT(Entropy) & 0.146$\pm$0.028 & 0.564$\pm$0.036   \\
    & BBT(MaxP) & 0.146$\pm$0.028 &0.568$\pm$0.043  \\
              & \textbf{Ours(ELBO)(Entropy)} & 0.132$\pm$0.009 & 0.542$\pm$0.006 \\
              & \textbf{Ours(ELBO)(MaxP)} & 0.132$\pm$0.009 & 0.540$\pm$0.009 \\
              & \textbf{Ours(Ensembles)(Entropy)}     & 0.104$\pm$0.014 & 0.525$\pm$0.023\\
              & \textbf{Ours(Ensembles)(MaxP)}     & 0.104$\pm$0.014 & 0.523$\pm$0.024 \\
    \hline \hline
    \makecell{Gradient-free \&}    & \textbf{Ours(ABC-SMC)(Entropy)} & 0.327$\pm$0.002 & 0.607$\pm$0.001 \\
    \makecell{Likelihood-free}& \textbf{Ours(ABC-SMC)(MaxP)} & 0.327$\pm$0.002 & 0.607$\pm$0.001 \\
    \toprule
  \end{tabular}
  \label{table:selective-backbone}
  \end{center}
  
\end{table}

\begin{table}[!t]
  \begin{center}
  \caption{\textbf{Far OOD Detection} (AURRRC score $\boldsymbol{\downarrow}$)}
  \vspace{-1.5em}
   \resizebox{\linewidth}{!}{\begin{tabular}{llll}
    \\
    \toprule
    \textbf{Settings} & \textbf{Methods} & \makecell[l]{ID:\textbf{SST-2}\\ OOD:\textbf{RTE}}
    &\makecell[l]{ID:\textbf{SNLI}\\ OOD:\textbf{MRPC}}\\
    \hline
    Gradient-free& BBT(entropy) & 0.402$\pm$0.015 & 0.076$\pm$0.016 \\
             & BBT(MaxP) & 0.402$\pm$0.015 & 0.072$\pm$0.015  \\
              & \textbf{Ours(ELBO)(Entropy)} & 0.365$\pm$0.037 & 0.089$\pm$0.007  \\
              & \textbf{Ours(ELBO)(MaxP)} & 0.365$\pm$0.037 & 0.084$\pm$0.006  \\
              & \textbf{Ours(Ensembles)(Entropy)}     & 0.338$\pm$0.027 & 0.074$\pm$0.020  \\
              & \textbf{Ours(Ensembles)(MaxP)}   & 0.338$\pm$0.027 & 0.071$\pm$0.018  \\
    \hline \hline
    \makecell{Gradient-free \&} & \textbf{Ours(ABC-SMC)(Entropy)} & 0.252$\pm$0.000 & 0.044$\pm$0.000 \\
    \makecell{Likelihood-free}& \textbf{Ours(ABC-SMC)(MaxP)} & 0.252$\pm$0.000  & 0.044$\pm$0.000 \\
    \toprule
  \end{tabular}}
  \label{table:far-OOD-detection-backbone}
  \end{center}
\end{table}

\subsection{Ablation of ABC-SMC sampling weights} \label{app:ablation-weights}
We compare both the prediction and uncertainty quantification performance of our proposed ABC-SMC approaches using the updated sampling weights and the fixed uniform weights. We denote the method using updated weights as ``ABC-SMC w. Weights". The results are shown in Table~\ref{table:test-acc-ablation}, \ref{table:calibration-ablation}, \ref{table:selective-ablation}, \ref{table:far-OOD-detection-ablation}, and \ref{table:near-OOD-detection-ablation}.
\begin{table*}[h]
  \begin{center}
  \caption{\textbf{Prediction Performance} (Test acc $\boldsymbol{\uparrow}$)}
  \vspace{-1.5em}
   \resizebox{\linewidth}{!}{\begin{tabular}{lllllllll}
    \\
    \toprule
    \textbf{Methods} & \textbf{SST-2} & \textbf{Yelp P.}& \textbf{AG’s News} & \textbf{DBPedia}& \textbf{MRPC} &\textbf{SNLI} &\textbf{RTE} &\textbf{Avg}\\
    \hline
    \textbf{ABC-SMC} & 86.51$\pm$0.55 & 90.32$\pm$0.03 & 81.43$\pm$0.41 & 57.41$\pm$0.90 & 80.78$\pm$0.07 & 40.81$\pm$0.24 & 53.37$\pm$0.30 & 70.09\\
    \textbf{ABC-SMC w. Weights} & 84.37$\pm$0.81 & 90.42$\pm$0.22 & 79.44$\pm$0.46 & 50.36$\pm$0.89 & 80.83$\pm$0.08 & 42.06$\pm$1.15 & 53.07$\pm$0.01 & 68.65\\
    \toprule
  \end{tabular}}
  \label{table:test-acc-ablation}
  \end{center}
  \vspace{-5em}
\end{table*}

\begin{table*}[h]
  \begin{center}
  \caption{\textbf{Calibration Performance} (ECE score $\boldsymbol{\downarrow}$)}
  \vspace{-1.5em}
   \resizebox{\linewidth}{!}{\begin{tabular}{lllllllll}
    \\
    \toprule
   \textbf{Methods} & \textbf{SST-2} & \textbf{Yelp P.}& \textbf{AG’s News} & \textbf{DBPedia}& \textbf{MRPC} &\textbf{SNLI} &\textbf{RTE} &\textbf{Avg}\\
    \hline
   \textbf{ABC-SMC} & 0.106$\pm$0.009 & 0.084$\pm$0.001 & 0.108$\pm$0.001 & 0.278$\pm$0.026 & 0.309$\pm$0.009 & 0.178$\pm$0.002 & 0.458$\pm$0.004 & 0.217\\
   \textbf{ABC-SMC w. Weights} & 0.156$\pm$0.008 & 0.091$\pm$0.005 & 0.160$\pm$0.023 & 0.506$\pm$0.009 & 0.316$\pm$0.002 & 0.182$\pm$0.005 & 0.463$\pm$0.003 & 0.268\\
    \toprule
  \end{tabular}}
  \label{table:calibration-ablation}
  \end{center}
   \vspace{-5em}
\end{table*}

\begin{table*}[h]
  \begin{center}
  \caption{\textbf{Selective Classification} (AURRRC score $\boldsymbol{\downarrow}$)}
  \vspace{-1.5em}
   \resizebox{\linewidth}{!}{\begin{tabular}{llllllllll}
    \\
    \toprule
    \textbf{Methods} & \textbf{SST-2} & \textbf{Yelp P.}& \textbf{AG’s News} & \textbf{DBPedia}& \textbf{MRPC} &\textbf{SNLI} &\textbf{RTE} &\textbf{Avg}\\
    \hline
    \textbf{ABC-SMC (Entropy)} & 0.061$\pm$0.006 & 0.075$\pm$0.002 & 0.110$\pm$0.004 & 0.285$\pm$0.015& 0.325$\pm$0.006  & 0.571$\pm$0.000 & 0.468$\pm$0.014 & 0.271\\
    \textbf{ABC-SMC(MaxP)} & 0.061$\pm$0.006 & 0.075$\pm$0.002 & 0.110$\pm$0.004 & 0.288$\pm$0.014& 0.325$\pm$0.006  & 0.579$\pm$0.000 & 0.468$\pm$0.014 & 0.272\\
    \textbf{ABC-SMC w. Weights (Entropy)} & 0.090$\pm$0.008 & 0.069$\pm$0.002 & 0.125$\pm$0.002 & 0.442$\pm$0.011& 0.315$\pm$0.001  & 0.570$\pm$0.019 & 0.460$\pm$0.006 & 0.296\\
    \textbf{ABC-SMC w. Weights (MaxP)} & 0.090$\pm$0.008 & 0.073$\pm$0.003 & 0.133$\pm$0.008 & 0.479$\pm$0.020 & 0.315$\pm$0.001  & 0.570$\pm$0.017 & 0.460$\pm$0.006 & 0.303\\
    \toprule
  \end{tabular}}
  \label{table:selective-ablation}
  \end{center}
   \vspace{-5em}
\end{table*}

\begin{table*}[h]
  \begin{center}
  \caption{\textbf{Far OOD Detection} (AURRRC score $\boldsymbol{\downarrow}$)}
  \vspace{-1.5em}
   \resizebox{\linewidth}{!}{\begin{tabular}{lllllllll}
    \\
    \toprule
   \textbf{Methods} & \makecell[l]{ID:\textbf{SST-2}\\ OOD:\textbf{RTE}} & \makecell[l]{ID:\textbf{Yelp P.}\\ OOD:\textbf{RTE}} &\makecell[l]{ID:\textbf{MRPC}\\ 
    OOD:\textbf{RTE}}&  \makecell[l]{ID:\textbf{DBPedia}\\ OOD:\textbf{AG’s News}}
    &\makecell[l]{ID:\textbf{SNLI}\\ OOD:\textbf{MRPC}}& \makecell[l]{ID:\textbf{RTE}\\ OOD:\textbf{MRPC}} &\textbf{Avg}\\
    \hline
    \textbf{ABC-SMC(Entropy)} & 0.126$\pm$0.009 & 0.005$\pm$0.001 & 0.396$\pm$0.001& 0.097$\pm$0.021 & 0.092$\pm$0.000 & 0.596$\pm$0.009 &0.219\\
    \textbf{ABC-SMC(MaxP)} & 0.126$\pm$0.009  & 0.005$\pm$0.001 & 0.396$\pm$0.001 & 0.095$\pm$0.021& 0.092$\pm$0.001 & 0.596$\pm$0.009 &0.218\\
    \textbf{ABC-SMC w. Weights(Entropy)} & 0.186$\pm$0.020 & 0.004$\pm$0.001 & 0.406$\pm$0.013& 0.079$\pm$0.013 & 0.067$\pm$0.004 & 0.596$\pm$0.006 &0.223\\
    \textbf{ABC-SMC w. Weights(MaxP)} & 0.186$\pm$0.020  & 0.004$\pm$0.001 & 0.402$\pm$0.006 & 0.091$\pm$0.002& 0.057$\pm$0.004 & 0.596$\pm$0.006 &0.223\\
    \toprule
  \end{tabular}}
  \label{table:far-OOD-detection-ablation}
  \end{center}
   \vspace{-5em}
\end{table*}

\begin{table*}[h]
  \begin{center}
  \small
  \caption{\textbf{Near OOD Detection} (AURRRC score $\boldsymbol{\downarrow}$)}
  \vspace{-1.5em}
  \begin{tabular}{lllllll}
    \\
    \toprule
    \textbf{Methods} & \makecell[l]{ID:\textbf{SST-2}\\ OOD:\textbf{IMDB}} & \makecell[l]{ID:\textbf{Yelp P.}\\ OOD:\textbf{IMDB}} &\makecell[l]{ID:\textbf{SNLI}\\ OOD:\textbf{MNLI}}& \makecell[l]{ID:\textbf{RTE}\\ OOD:\textbf{MNLI}} &\textbf{Avg}\\
    \hline
    \textbf{ABC-SMC(Entropy)} & 0.983$\pm$0.001 & 0.365$\pm$0.001 & 0.710$\pm$0.002 & 0.952$\pm$0.000 & 0.753 \\
    \textbf{ABC-SMC(MaxP)} & 0.983$\pm$0.001 & 0.365$\pm$0.001 & 0.694$\pm$0.000 & 0.952$\pm$0.000 & 0.749 \\
    \textbf{ABC-SMC w. Weights(Entropy)} & 0.967$\pm$0.002 & 0.365$\pm$0.001 & 0.572$\pm$0.049 & 0.952$\pm$0.001 & 0.714 \\
    \textbf{ABC-SMC W. Weights(MaxP)} & 0.967$\pm$0.002 & 0.365$\pm$0.001 & 0.534$\pm$0.032 & 0.952$\pm$0.001 & 0.705 \\
    \toprule
  \end{tabular}
  \label{table:near-OOD-detection-ablation}
  \end{center}
   \vspace{-5em}
\end{table*}